\documentclass{article}
\usepackage{ijcai25}

\usepackage{times}
\usepackage{soul}
\usepackage{url}
\usepackage[hidelinks]{hyperref}
\usepackage[utf8]{inputenc}
\usepackage[small]{caption}
\usepackage{graphicx}
\usepackage{amsmath}
\usepackage{amsthm}
\usepackage{booktabs}
\usepackage{algorithm}
\usepackage{algorithmic}
\usepackage[switch]{lineno}
\usepackage{multirow}
\usepackage{amssymb}
\usepackage{subcaption}

\theoremstyle{definition}
\newtheorem{definition}{Definition}

\title{ADFormer: Aggregation Differential Transformer for Passenger Demand Forecasting}

\author{
Haichen Wang$^1$\and
Liu Yang$^1$\and
Xinyuan Zhang$^1$\and
Haomin Yu$^2$\and
Ming Li$^3$\And
Jilin Hu$^{1,4,\ast}$\\
\affiliations
$^1$East China Normal University\\
$^2$Aalborg University\\
$^3$INSPUR Co.,Ltd\\
$^4$KLATASDS-MOE
\emails
\{hcwang, lyang, xyZhang\_29\}@stu.ecnu.edu.cn,
haominyu@cs.aau.dk,
liming2017@inspur.com,
jlhu@dase.ecnu.edu.cn
}
\begin{document}
\maketitle
\renewcommand{\thefootnote}{\fnsymbol{footnote}}
\footnotetext[1]{Corresponding author: jlhu@dase.ecnu.edu.cn}
\begin{abstract}
Passenger demand forecasting helps optimize vehicle scheduling, thereby improving urban efficiency. Recently, attention-based methods have been used to adequately capture the dynamic nature of spatio-temporal data. However, existing methods that rely on heuristic masking strategies cannot fully adapt to the complex spatio-temporal correlations, hindering the model from focusing on the right context. These works also overlook the high-level correlations that exist in the real world. Effectively integrating these high-level correlations with the original correlations is crucial. To fill this gap, we propose the Aggregation Differential Transformer (ADFormer), which offers new insights to demand forecasting promotion. Specifically, we utilize Differential Attention to capture the original spatial correlations and achieve attention denoising. Meanwhile, we design distinct aggregation strategies based on the nature of space and time. Then, the original correlations are unified with the high-level correlations, enabling the model to capture holistic spatio-temporal relations. Experiments conducted on taxi and bike datasets confirm the effectiveness and efficiency of our model, demonstrating its practical value. The code is available at \url{https://github.com/decisionintelligence/ADFormer}.
\end{abstract}

\section{Introduction}
\begin{figure}
    \centering
    \includegraphics[width=\linewidth]{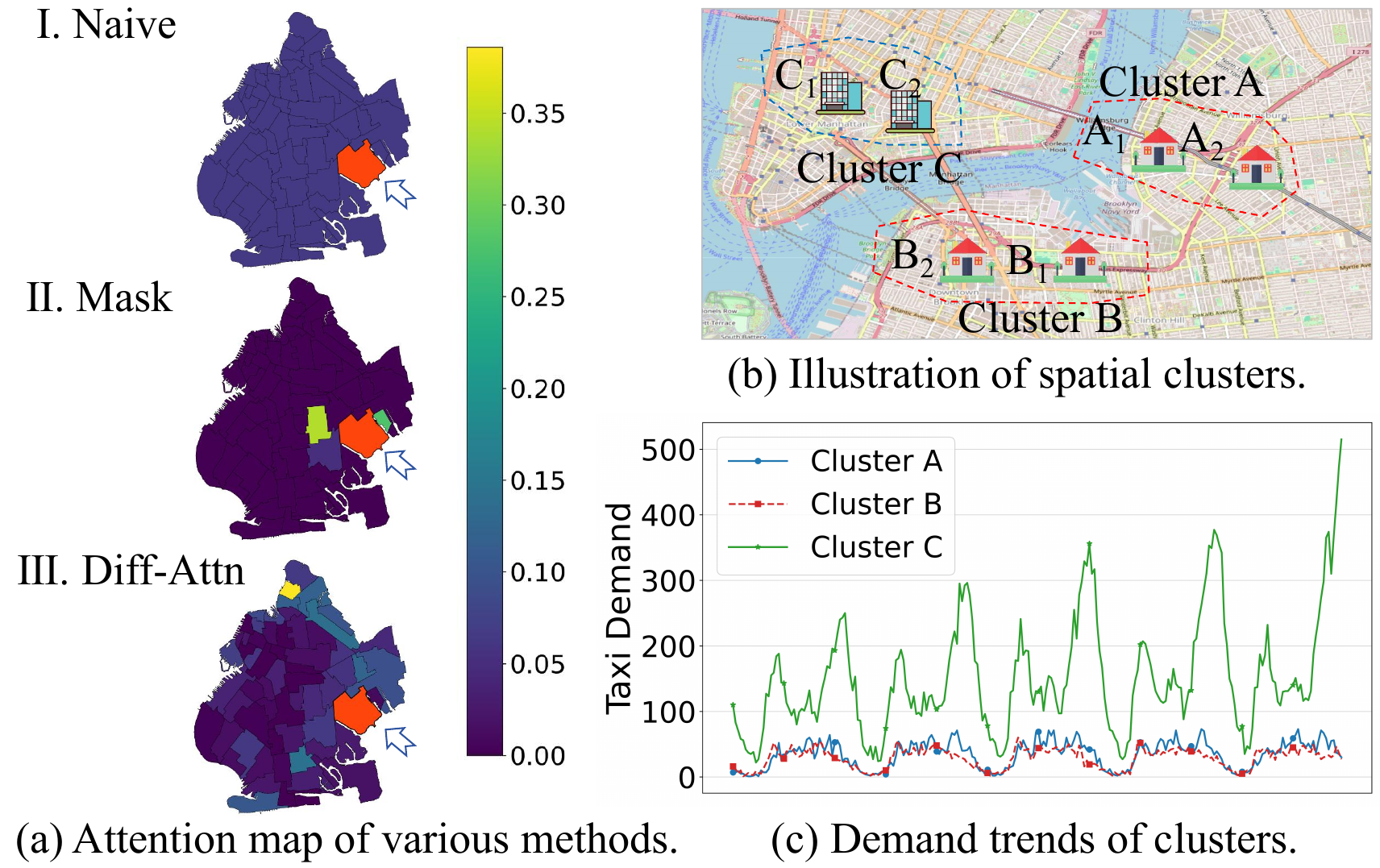}
    \caption{Illustration of spatial attention noise and high-level correlations.
    }
    \vspace{-0.5cm}
    \label{fig:st-agg}
\end{figure}
With the rise of digital and intelligent technologies, travel is becoming increasingly smart. Ride-hailing and bike-sharing services reflect this trend~\cite{chen2018price}, but face inefficiencies like long wait times, vehicle idling, and unbalanced bicycle distribution. Accurate predictions are of great importance in real-world applications.~\cite{DBLP:journals/corr/abs-2307-00495}~\cite{qiu2024tfb}~\cite{li2024foundts}~\cite{di2024machine}

The key challenge in spatio-temporal data modeling lies in determining the appropriate structure to capture intricate spatio-temporal dependencies.
The topological relationships of cities describe the adjacency or containment relationships between regions, making spatio-temporal forecasting well suited for modeling with graph neural networks~(GNNs). Consequently, GNNs have been widely applied in this field, such as~\cite{geng2019spatiotemporal}~\cite{hu2020stochastic}~\cite{zheng2021soup}. 
Next, with the success of attention models, some works argue that despite the notable success of GNN-based approaches, defining the adjacency matrix is challenging and they struggle to capture the dynamic nature of spatio-temporal data~\cite{yan2021learning}. Therefore, recent studies~\cite{jiang2023pdformer}~\cite{lu2024stpsformer}~\cite{liang2024asstformer} adopted attention-based methods to address spatio-temporal problems, which can sufficiently understand the dynamics of spatial and temporal correlations. \textbf{However, the attention-based methods still suffer from the following drawbacks despite their effectiveness. }

\textit{Firstly, the attention-based method tends to bring in attention noise that may disproportionately focus on irrelevant regions.} As illustrated in~\autoref{fig:st-agg}(a), we visualize the correlations between a specific region (in red and pointed by the arrows) and others under various methods. The observations are: 1) The naive attention mechanism exhibits dispersed relationships with generally low correlations across different regions. 2) The mask-based mechanism focuses only on local regions and few relevant areas. However, it is beneficial for us to concentrate on critical information, while most other areas maintain low correlations, as the bottom map depicts in~\autoref{fig:st-agg}(a). 

\textit{Secondly, current methodologies fail to adequately account for higher-level correlations from both spatial and temporal perspectives. }For example, in~\autoref{fig:st-agg}(b), the map depicts six distinct regions, where regions $A_1$ and $A_2$, as well as $B_1$ and $B_2$, represent residential areas, while $C_1$ and $C_2$ correspond to commercial areas. It is intuitive that regions in close geographical proximity exhibit similar dynamic patterns. 
Then, we aggregate regions $A_1$ and $A_2$ into a single entity, termed \textit{cluster $A$}, and group $B_1$ and $B_2$ into \textit{cluster $B$}, $C_1$ and $C_2$ into \textit{cluster $C$}, we observe from~\autoref{fig:st-agg}(c) that \textit{cluster $A$} and \textit{cluster $B$} exhibit similar demand patterns, while \textit{cluster $C$} demonstrates distinct behavior. Thus, it is vital to consider higher-level correlations from the spatial perspective. 

Further, strong temporal correlations must also be considered. For instance, taxi demand is typically high in residential areas and low in commercial areas during morning peak hours, while the reverse pattern occurs in afternoon peak hours. Furthermore, factors such as varying work schedules across companies lead to dynamic demand correlations over a broader time span. Existing approaches, such as those proposed by~\cite{guo2023self} and~\cite{huang2025std}, incorporate reference point learning within attention mechanisms to capture high-order correlations. However, these methods are failing to adequately explore high-level temporal information. Additionally, they struggle to effectively integrate correlations from all sources, posing a challenge in accurate passenger demand forecasting.

 In this paper, we propose the Aggregation Differential Transformer (ADFormer) to effectively address the aforementioned challenges. To capture meaningful high-level spatial correlations, we introduce a \textit{Unified Spatial Attention} module, which integrates Differential Attention to enhance spatial correlation detection while suppressing attention noise. We further apply spatial aggregation to identify and model high-level regional dependencies. For temporal modeling, we present a \textit{Systemic Temporal Attention} module, consisting of temporal aggregation attention and self-attention blocks. The temporal aggregation employs a hierarchical temporal matrix, reconstructed via a learnable mapping from temporal information. In summary, the contributions of this paper are as follows:


\begin{itemize}
    \item To reduce attention noise in spatial correlation detection, we apply Differential Attention to refine the attention matrix and enhance its robustness.
    \item To capture comprehensive spatio-temporal dependencies, we propose a region aggregation strategy that models non-pairwise spatial relationships while preserving original spatial correlations, along with a hierarchical temporal matrix to mitigate disruptions in temporal continuity caused by irregular time intervals.

    \item Extensive experiments conducted on three real-world datasets from two cities demonstrate that our proposed model, ADFormer, surpasses state-of-the-art baselines in forecasting accuracy while maintaining computational efficiency, highlighting its practical applicability.
\end{itemize}

\section{Preliminaries}
\subsection{Notations and Definitions}
\begin{definition}[Urban Region]
An urban region corresponds to a boundary, such as a zone, within the real world context, denoted as $r$. The area of interest can be partitioned into a set of non-overlapping regions, denoted as $\mathcal{R} = \{r_1, r_2, \dots, r_N\}$, where $N$ is the total number of urban regions. 
\end{definition}

\begin{definition}[Spatial Cluster]
A spatial cluster is defined as a set of urban regions aggregated based on a specific aggregation criterion, denoted as $C = \{r_1, r_2, \dots, r_{|C|}\}$. For instance, if the aggregation criterion is functional similarity, the regions within a cluster share similar functionalities. Under this criterion, the area of interest can be partitioned into a set of disjoint clusters, represented as $\mathcal{C} = \{C_1, C_2, \dots, C_{M}\}$, where $M$ is the total number of clusters.
\end{definition}

\begin{definition}[Region Passenger Demand]
Passenger demand is defined as the total number of passengers requesting service within a region at a specific time step $t$, denoted as $x_t \in \mathbb{R}^{D}$, where $D$ represents the dimensionality of the features. The passenger demand across an urban region over the past $T$ time steps is represented as $X^{raw} \in \mathbb{R}^{T \times N \times D}$. 
\end{definition}


\subsection{Problem Statement}
\begin{figure*}
    \centering
    \includegraphics[width=\linewidth]{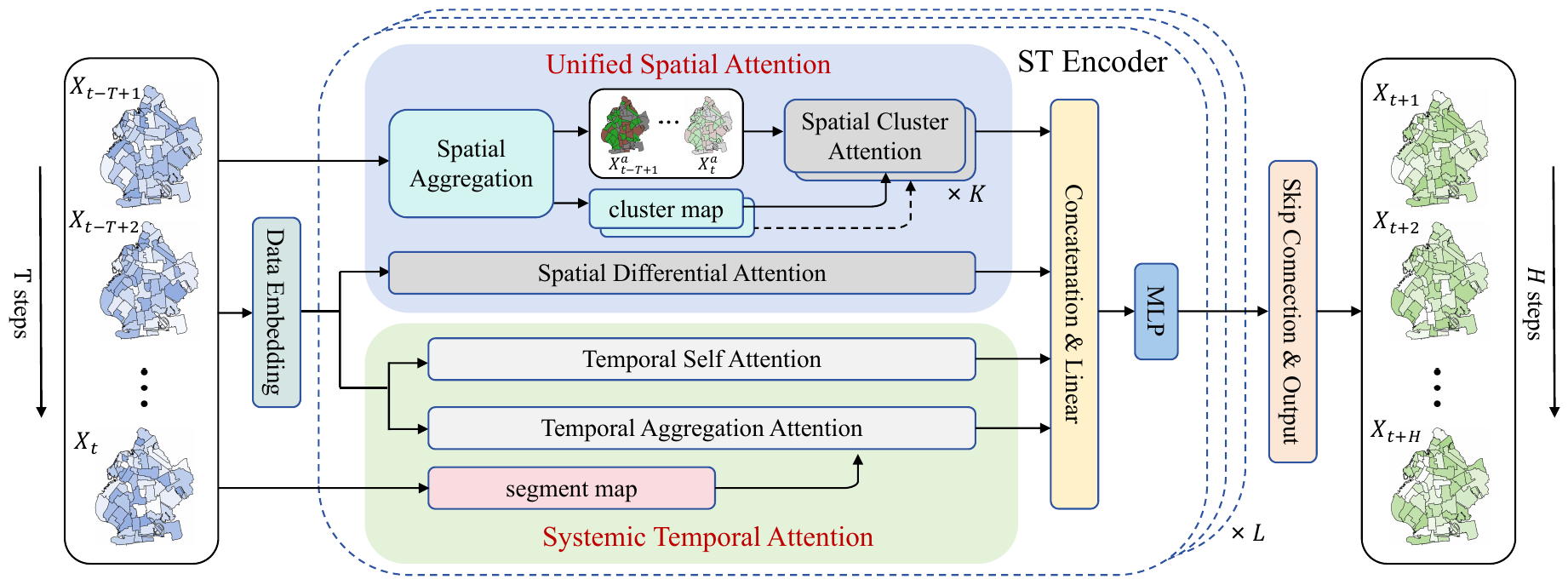}
    \caption{The overall framework of ADFormer.}
    \label{fig:model}
\end{figure*}
Given historical observation data $[X^{raw}_{t-T+1}, \dots, X^{raw}_{t}]$, our goal is to learn a mapping function $f(\cdot)$ that can generate accurate predictions for future demand, which can be formulated as follows. 
\begin{equation}
    \resizebox{0.91\linewidth}{!}{$
        [X^{raw}_{t-T+1},\dots,X^{raw}_{t};\mathcal{R}]\xrightarrow{f(\cdot)}[X_{t+1},\dots,X_{t+H}],
    $}
\end{equation}
where $T$ represents the historical window and $H$ denotes the future horizon. 

\section{Methodology}
In this section, we first introduce how to embed the data into the high-dimensional space. Then, we explain how to capture their original correlations and high-level correlations from spatial and temporal perspectives, respectively. Finally, we describe the way to fuse them in the output layer. The overall framework is presented in \autoref{fig:model}.

\subsection{Data Embedding}
To improve the model's ability to capture the periodic patterns inherent in passenger demand, we augment the original data with time-related features. These features include the \textit{time of day} and the \textit{day of the week}, which reflect the significant periodicity in passenger demand:
\begin{equation}
    X_{full} = X_{raw} \mathbin{\|} T_d \mathbin{\|} T_w,
\end{equation}
where $\mathbin{\|}$ denotes the concatenation operator and $X_{full}\in \mathbb{R}^{T \times N \times D'}$. $T_d \in \mathbb{R}^{T \times N \times 1}$ represents the time of day and $T_w \in \mathbb{R}^{T \times N \times 7}$ is the one-hot vector of the day of week.

Then, fully connected layers are used to map $X_{raw}$ and the additional time-related information to high-dimensional space separately, and subsequently sum them together:
\begin{equation}
    X_{emb} = X_{raw}W_{raw} + T_dW_{T_d} + T_wW_{T_w},
\end{equation}
where $X_{emb} \in \mathbb{R}^{T \times N \times d}$. $W_{raw} \in \mathbb{R}^{D \times d}, W_{T_d} \in \mathbb{R}^{1 \times d}$ and $W_{T_w} \in \mathbb{R}^{7 \times d}$ are learnable parameters.

Since the self-attention mechanism does not inherently contain sequential information, we adopt the classical positional encoding scheme, leveraging the periodicity of sine and cosine functions to generate positional information for the input sequence, namely $X_{pos} \in \mathbb{R}^{T \times d}$. 

Furthermore, regions have distinct roles in a city and their passenger demand patterns vary accordingly. Inspired by~\cite{shao2022spatial}, a simple yet effective model, we introduce a spatial embedding $X_{spa} \in \mathbb{R}^{N \times d}$ as an identifier for each region. Finally, all components are summed and used as the subsequent input:
\begin{equation}
    X_{input} = X_{emb} + X_{pos} + X_{spa}.
\end{equation}
$X$ will be used to replace $X_{input}$ for convenience in the following text.

\subsection{Unified Spatial Attention}

To consider spatial correlations of both urban regions and spatial clusters, we propose a unified spatial attention module that consists of \textit{1)~Spatial Differential Attention} and \textit{2)~Spatial Cluster Attention}, which are detailed as follows. 

\subsubsection{Spatial Differential Attention}

To address the issue of attention noise in spatial correlations at the urban region level, we examine the limitations of the original attention mechanism. It has been demonstrated that the original attention mechanism tends to over-attend to irrelevant context, a phenomenon that is particularly pronounced in spatial attention applications, especially when considering urban regions where a larger number of elements are involved in the attention computation. To mitigate this issue, we apply the recently proposed differential attention mechanism, as introduced in large language model studies~\cite{ye2024differential}, to the spatio-temporal data domain. Following the structure of the differential attention mechanism, we first compute the Query, Key, and Value matrices as outlined below:

\begin{equation}
    Q_1^S\mathbin{\|}Q_2^S = XW^S_Q,\ K_1^S\mathbin{\|}K_2^S = XW^S_K,\ V^S = XW^S_V,
\end{equation}
where $W^S_Q, W^S_K, W^S_V\in \mathbb{R}^{d \times d^s_1}$ are learnable parameters. $Q_1^S, Q_2^S, K_1^S, K_2^S\in \mathbb{R}^{T \times N \times (d^s_1/2)}$ are the queries and keys of the computation of two attention score matrices. $V^S\in \mathbb{R}^{T \times N \times d^s_1}$ is the value of the differential attention operation. 

Then, we can obtain the spatial output at the urban region level based on the differential attention as follows: 
\begin{align}
    \text{SDA}(X) = & (\text{softmax}(\frac{Q_1^S(K_1^S)^T}{\sqrt{d^s_1/2}}) - \\ 
    & \nonumber \lambda \ \text{softmax}(\frac{Q_2^S(K_2^S)^T}{\sqrt{d^s_1/2}}))V^S,
\end{align}
\noindent
where $\lambda$ is a learnable scalar. The result of the difference of the two attention matrices represents the denoised attention scores among the $N$ regions. Aligned with~\cite{ye2024differential}, we set the scalar $\lambda$ as:
\begin{equation}
    \lambda = \text{exp} (\lambda_{q_1} \cdot \lambda_{k_1}) - \text{exp} (\lambda_{q_2} \cdot \lambda_{k_2}) + \lambda_{\text{init}},
\end{equation}
where $\lambda_{q_1}, \lambda_{k_1}, \lambda_{q_2}$ and $\lambda_{k_2} \in \mathbb{R}^{d^s_1/2}$ are learnable vectors. $\lambda_{\text{init}}$ is a randomly initialized constant.


\subsubsection{Spatial Cluster Attention}
To capture high-level correlations among regions, we aggregate them based on the similarity of their historical demand data, which reflects their functional roles within the urban environment. This approach groups regions with similar demand patterns, even if they are geographically distant. For instance, two commercial centers located far apart may exhibit comparable trends due to similar functions. Such aggregation enables the model to capture functional similarities beyond spatial proximity.
 
Specifically, we calculate the similarity of historical data between two regions through the Dynamic Time Warping (DTW)~\cite{berndt1994using} algorithm, further attaining the distance matrix $M_{sim} \in \mathbb{R}^{N \times N}$, which describes the similarity distances between regions.

Regions are aggregated based on $M_{sim}$. Initially, each region is considered as a separate cluster. Then, calculate distances between clusters:
\begin{equation}
    D(C_i, C_j) = \frac{1}{|C_i|\cdot|C_j|}\sum_{p\in C_i}\sum_{q\in C_j}M_{sim}(p,q),
\end{equation}
where $C_i=\{r_1,\dots,r_{|C_i|}\}$ and $C_j=\{r_{1'},\dots,r_{|C_j|}\}$. Iteratively merge the closest clusters until the desired number of clusters is achieved, resulting in the final clustering $\mathcal{C}$.

For the aggregated clusters, we adjust the cluster sizes to ensure that the aggregation results are balanced to some extent. For instance, move region $p$ to $C_j$ if the size of the cluster to which $p$ belongs exceeds the threshold and the distance to the cluster $C_j$ is the smallest. 

In this way, we can identify which original regions are included in the clusters, helping us to get the cluster map, denoted as $M_{cls} \in \mathbb{R}^{M\times N}$. Based on this, we aggregate the $X_{raw}$ to obtain $X_{agg}$:
\begin{equation}
    X_{agg} = M_{cls}X_{raw}\mathbin{\|}T_d\mathbin{\|}T_w.
\end{equation}
$X_{agg}$ eventually becomes $X_{agg}^{emb} \in \mathbb{R}^{T \times M \times d}$ via Data Embedding with temporal information enhancement and spatial identity assignment. We will use $X_{a}$ to replace $X_{agg}^{emb}$ for convenience. Next, we employ spatial cluster self-attention (SCA) to model the high-level correlations:
\begin{align}
    &Q^C = X_aW^{C}_Q, \ K^C = X_aW^{C}_K, \ V^C = X_aW^{C}_V, \\ &\text{SCA}(X_a) = \text{softmax}((M^S_{sep})^T\frac{Q^C(K^C)^T}{\sqrt{d^s_2}})V^C,
\end{align}
where $W^{C}_Q, W^{C}_K, W^{C}_V \in \mathbb{R}^{d \times d^s_2}$ are learnable mapping matrices. After capturing the high-level correlations, we need to map these back to the original spatial level to assist in forecasting the demand volume in the regions. We take advantage of the $M_{cls}$ to help initialize the learnable separation parameter, aiming to allocate the correlations of clusters to regions that are closely related to them:
\begin{equation}
    M_{sep}^S = M_{cls}\odot M_{sep},
\end{equation}
where $\odot$ represents Hadamard product and $M_{sep} \in \mathbb{R}^{M \times N}$ is a randomly initialized parameter.

\subsection{Systemic Temporal Attention}
\begin{figure}
    \centering
    \includegraphics[width=\linewidth]{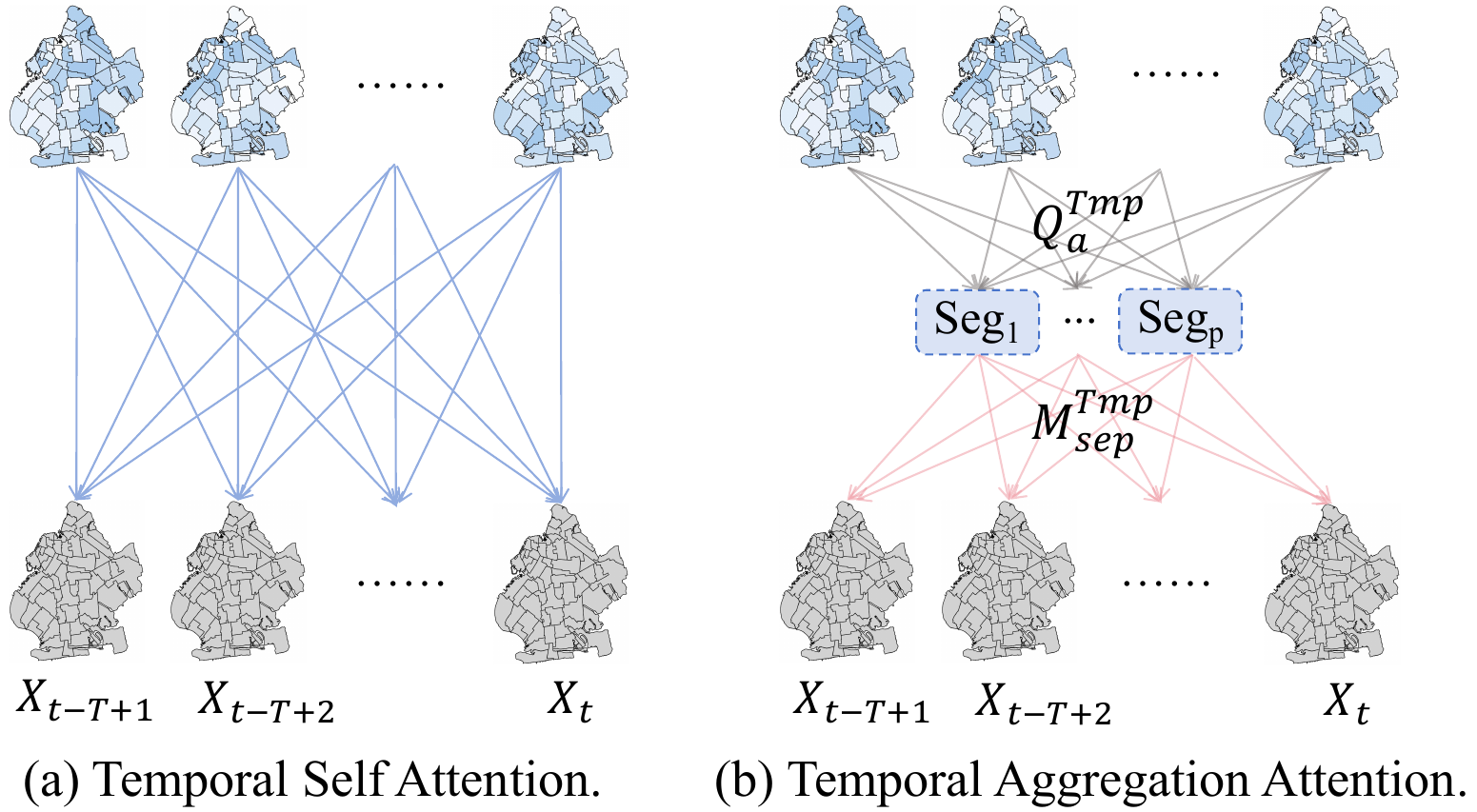}
    \caption{Systemic Temporal Attention.}
    \label{fig:tmp-agg}
\end{figure}
Temporal correlation is another important aspect of passenger demand forecasting, which also involves multi-level correlations. Thus, we propose a Systemic Temporal Attention module, which is shown in~\autoref{fig:tmp-agg}. This module consists of a)~\textit{Temporal Self Attention} and b)~\textit{Temporal Aggregation Attention} to capture the multi-level temporal correlations. It is worth mentioning that, due to the relatively few time steps in passenger demand forecasting, we do not use Differential Attention to denoise attention along the temporal dimension.

\subsubsection{Temporal Self Attention}
We make use of the self-attention mechanism to extract temporal correlations among regions. Firstly, we transpose $X$ to $X^T \in \mathbb{R}^{N \times T \times d}$ and map it to query, key and value:
\begin{equation}
    \resizebox{0.89\linewidth}{!}{$
        Q^{Tmp} = X^TW^{Tmp}_Q, \ K^{Tmp} = X^TW^{Tmp}_K, \ V^t = X^TW^{Tmp}_V,
    $}
\end{equation}
where $W^{Tmp}_Q, W^{Tmp}_K, W^{Tmp}_V \in \mathbb{R}^{d \times d^t_1}$ are learnable mapping parameters.

Then, we compute the attention scores to obtain the correlations between time steps, thereby deriving the output of temporal self attention:
\begin{equation}
    \text{TSA}(X^T) = \text{softmax}(A^{Tmp})V^{Tmp},
\end{equation}
where $A^{Tmp}  = \frac{Q^{Tmp}(K^{Tmp})^T}{\sqrt{d^t_1}} \in \mathbb{R}^{N \times T \times T}$ represents the correlations between steps among regions.

\subsubsection{Temporal Aggregation Attention}
Similar to spatial aggregation, it is also difficult to obtain an explicit aggregation in passenger demand forecasting. In this paper, we employ a learnable matrix $Q_a^{Tmp} \in \mathbb{R}^{N \times P \times d^t_2}$ to act as the query during the calculation of attention:
\begin{align}
    &K^{Tmp}_a = X^TW^{Tmp_{a}}_K, \ V^{Tmp}_a = X^TW^{Tmp_{a}}_V, \\
    &\resizebox{0.89\linewidth}{!}{$
        \text{TAA}(X^T) = \text{softmax}(M^{Tmp}_{sep}\frac{Q^{Tmp}_a(K^{Tmp}_a)^T}{\sqrt{d^t_2}})V^{Tmp}_a,
    $}
\end{align}
where $W^{Tmp_{a}}_K, W^{Tmp_{a}}_V \in \mathbb{R}^{d \times d^t_2}$ are mapping parameters.
We leverage temporal information within the data to derive the restoration matrix $M^{Tmp}_{sep}\in\mathbb{R}^{N\times T\times P}$:
\begin{equation}
    M_{sep}^{Tmp} = (X_{full}[:,:,D:D'])^TW_{sep},
\end{equation}
where $(X_{full}[:,:,D:D'])^T \in \mathbb{R}^{N \times T \times (D'-D)}$ and $W_{sep} \in \mathbb{R}^{(D'-D) \times P}$ is a learnable parameter.


\subsection{Output Layer}
\subsubsection{Attention Fusion} We need to integrate the attention mechanisms applied in multiple aspects.
For spatial dimension, we perform multilevel spatial aggregations in the encoder of each layer, namely $\mathcal{X}_a=\{ X_{a_1},\dots,X_{a_K} \}$. Accordingly, $K$ calculations of SCA($X_{a_i}$) are required. Overall, spatial attention (SA) can be formally expressed as:
\begin{equation}
    \text{SA}(X, \mathcal{X}_a) = \text{SDA}(X) \mathbin{\|} \sum_{i=1}^{K} \text{SCA}(X_{a_i}),
\end{equation}
where $\mathbin{\|}$ denotes concatenation and each aggregation level has an associated cluster map in spatial cluster attention. 

In terms of temporal dimension, we use only one aggregation per layer to reduce complexity. Temporal attention (TA) can be expressed in a formulated way:
\begin{equation}
    \text{TA}(X) = \text{TSA}(X^T) \mathbin{\|} \text{TAA}(X^T).
\end{equation}

\subsubsection{Spatial-Temporal Encoder}
After merging the spatial and temporal attention at each level respectively, we integrate all components globally and perform forward propagation via fully connected layers:
\begin{align}
    &\text{Encoder}(X, \mathcal{X}_a) = \text{LayerNorm}(\text{MLP}(H) + H), \\
    &H = \text{LayerNorm} ((\text{SA}(X, \mathcal{X}_a) \mathbin{\|} \text{TA}(X))W^O + X),
\end{align}
where $W^O \in \mathbb{R}^{(d_1^s + d_2^s + d^t_1 + d^t_2) \times d}$ is a learnable projection parameter.
We stack $L$ encoder layers to obtain the final output, and utilize layer normalization~\cite{ba2016layer} and residual connection~\cite{he2016deep} to stabilize the training process.

\section{Experiments}
\begin{table*}

    \centering
    \renewcommand{\arraystretch}{0.6}
    \resizebox{2.0\columnwidth}{!}
    {
    \tiny
    \begin{tabular}{c|c|l|rrrrrrrr}
        \toprule
        Dataset & Horizon & Metric & GWNet & MTGNN & AGCRN & RGSL & GMAN & PDFormer & ASSTFormer & ADFormer \\ 
        \midrule
        \multirow{9}{*}{\raisebox{-3.5ex}[0pt][0pt]{\shortstack{NYC\\Taxi}}}
            & \multirow{3}{*}{30 min} & MAE    & 5.697  & 5.677  & 5.797  & 5.724  & 5.662  & 5.625 & 5.756    & \textbf{5.461} \\
            &                         & RMSE   & 11.871 & 11.833 & 13.304 & 12.555 & 11.750 & 11.597 & 11.953  & \textbf{11.342} \\
            &                         & MAPE   & 18.919 & 18.718 & 18.768 & 18.677 & 19.081 & 18.920 & 19.185  & \textbf{18.238} \\ \cmidrule{2-11}
            & \multirow{3}{*}{90 min} & MAE    & 6.344  & 6.229  & 6.486  & 6.388  & 6.273  & 6.131 & 6.358   & \textbf{5.975} \\
            &                           & RMSE   & 13.687 & 13.532 & 14.600 & 14.327 & 13.565 & 13.128 & 13.807  & \textbf{12.920} \\
            &                           & MAPE   & 20.237 & \textbf{19.758} & 20.171 & 20.027 & 20.134 & 21.036 &21.422  & 20.348 \\ \cmidrule{2-11}
            & \multirow{3}{*}{3 hour} & MAE    & 6.922  & 6.809  & 6.979  & 7.052  & 6.897  & 6.684 & 7.053   & \textbf{6.543} \\
            &                         & RMSE   & 15.363 & 15.158 & 16.222 & 16.133 & 15.397 & 14.776 & 16.135  & \textbf{14.592} \\ 
            &                         & MAPE   & 21.303 & 21.162 & 21.406 & 21.704 & 21.785 & 21.129 & 22.244  & \textbf{20.619} \\
        \midrule
        \multirow{9}{*}{\raisebox{-3.5ex}[0pt][0pt]{\shortstack{NYC\\Bike}}}
 
            & \multirow{3}{*}{30 min} & MAE    & 3.493  & 3.405  & 3.401  & 3.371  & 3.416  & 3.397 & 3.463   & \textbf{3.299} \\
            &                         & RMSE   & 5.569  & 5.348  & 5.381  & 5.289  & 5.321  & 5.230 & 5.409   & \textbf{5.104} \\
            &                         & MAPE   & 25.531 & 25.009 & 25.044 & 24.977 & 25.323 & 25.251 & 25.808  & \textbf{24.585} \\ \cmidrule{2-11}
            & \multirow{3}{*}{90 min} & MAE    & 3.899  & 3.731  & 3.727  & 3.674  & 3.765  & 3.741 & 3.799   & \textbf{3.588} \\
            &                           & RMSE   & 6.433  & 6.016  & 6.059  & 5.943  & 6.073  & 5.973  & 6.211  & \textbf{5.731} \\
            &                           & MAPE   & 28.529 & 27.381 & 27.657 & 27.248 & 27.858 & 27.499 & 28.084  & \textbf{26.664} \\ \cmidrule{2-11}
            & \multirow{3}{*}{3 hour} & MAE    & 4.481  & 4.179  & 4.212  & 4.176  & 4.271  & 4.215 & 4.167   & \textbf{4.029} \\
            &                         & RMSE   & 7.685  & 7.011  & 7.029  & 7.065  & 7.227  & 7.107 & 7.038   & \textbf{6.810} \\ 
            &                         & MAPE   & 32.909 & 31.856 & 32.528 & 32.054 & 31.964 & 32.635 & 32.754  & \textbf{31.563} \\
        \midrule
        \multirow{9}{*}{\raisebox{-3.5ex}[0pt][0pt]{\shortstack{Xi'an\\Taxi}}}

            & \multirow{3}{*}{30 min} & MAE    & 3.671  & 3.623  & 3.684  & 3.702  & 3.812  & 3.741 & 3.696    & \textbf{3.564} \\
            &                         & RMSE   & 5.357 & 5.312 & 5.553 & 5.503 & 5.576 & 5.453 & 5.415  & \textbf{5.208} \\
            &                         & MAPE   & 22.631 & \textbf{22.159} & 22.353 & 22.502 & 24.037 & 24.098 & 23.487  & 22.848 \\ \cmidrule{2-11}
            & \multirow{3}{*}{90 min} & MAE    & 3.852  & 3.804  & 3.843  & 3.845 & 3.973 & 3.887 & 3.865  & \textbf{3.697} \\
            &                           & RMSE   & 5.693 & 5.663 & 5.794 & 5.768 & 5.902 & 5.727 & 5.735  & \textbf{5.466} \\
            &                           & MAPE   & 23.504 & 23.107 & 23.168 & 23.195 & 24.216 & 23.660 & 23.232  & \textbf{22.464} \\ \cmidrule{2-11}
            & \multirow{3}{*}{3 hour} & MAE    & 4.011  & 3.959  & 3.961 & 4.046 & 4.146  & 4.074 & 4.034   & \textbf{3.830} \\
            &                         & RMSE   & 6.003 & 5.959 & 5.976 & 6.159 & 6.228 & 6.061 & 6.061  & \textbf{5.706} \\ 
            &                         & MAPE   & 24.229 & 23.778 & 23.822 & 24.168 & 24.723 & 24.589 & 23.826  & \textbf{23.114} \\
        \bottomrule
    \end{tabular}
    }
    \caption{Performance comparison of different methods on public demand datasets.}
    \label{tab:performance_comparison}
    \vspace{-0.2cm}
\end{table*}
\subsection{Experimental Settings}

\subsubsection{Datasets} We evaluate our model on three widely used public datasets:
{\renewcommand{\thefootnote}{\arabic{footnote}}
NYC-Taxi\footnote[1]{https://www.nyc.gov/site/tlc/about/tlc-trip-record-data.page}},
{\renewcommand{\thefootnote}{\arabic{footnote}}
NYC-Bike\footnote[2]{https://citibikenyc.com/system-data}}, and Xi'an-Taxi, which exhibit diverse urban structures and demand distributions.
\textbf{NYC-Taxi/Bike.} These two datasets record transportation activity across 263 boroughs (i.e., urban regions) in New York. Each taxi or bike trip includes information such as start/end time, locations, cost, and other details. Based on pickup and drop-off times, we record the number of passengers or rides in 30-minute intervals. The NYC-Taxi data spans January–December 2016, while the NYC-Bike data covers the same period in 2023.
\textbf{Xi'an-Taxi.} This dataset, provided by Didi, a ride-hailing company in China, contains vehicle trajectories during passenger trips. Each trajectory includes timestamps and GPS coordinates. We extract trip origin points and divide the city into non-overlapping hexagonal grids to generate demand sequences. The data spans October and November 2016.

For each dataset, we split the data into training, validation, and test sets in a 7:1:2 ratio. For single-step prediction, the past three hours (6 steps) of demand data are used to forecast the next 30 minutes (1 step). For multi-step prediction, the same input is used to predict demand over the next 1.5 hours (3 steps) and 3 hours (6 steps).

\subsubsection{Baselines} We compare ADFormer with seven baselines across two categories. 
(1) GCN-based models. 
GWNet~\cite{wu2019graph}, 
MTGNN~\cite{wu2020connecting}, 
AGCRN~\cite{bai2020adaptive}, 
RGSL~\cite{RGSLyu}.
(2) Attention-based models. 
GMAN~\cite{zheng2020gman}, 
PDFormer~\cite{jiang2023pdformer}, 
ASSTFormer~\cite{liang2024asstformer}. 

\subsubsection{Model Parameters}
The experimental environment is configured as follows: the Python version is 3.10.0, the CUDA version is 12.1, and the PyTorch version is 2.5.1. We conduct experiments on an NVIDIA GeForce RTX 3090 with 24GB of memory. The AdamW optimizer is used in model training, with an initial learning rate of 1e-3, decaying to 1e-4. We explore hidden dimension $\in {\{32, 64, 128\}}$, the depth of encoder $\in {\{4, 6, 8\}}$ and impact of number of spatial clusters and hierarchical levels in the \textit{Parameter Study}.

\subsubsection{Metrics} We take Mean Absolute Error~(MAE), Root Mean Squared Error~(RMSE) and Mean Absolute Percentage Error~(MAPE) as evaluation metrics. 
Following~\cite{yao2018deep}, we exclude steps with low demand and set the threshold to 5, which means that points less than the value are not considered.


\subsection{Overall Performance}
As shown in~\autoref{tab:performance_comparison}, our model consistently outperforms baselines, demonstrating strong performance in real-world demand forecasting. Among dynamic graph methods, the performance is similar due to shared message-passing mechanisms. MTGNN stands out by jointly modeling time series and learning graph structures.

Attention-based methods generally outperform graph-based ones by better capturing dynamic correlations. PDFormer stands out for its ability to model spatio-temporal dynamics during attention computation and filter out low-relevance pairs, leading to improved performance. ASSTFormer performs well on datasets with fewer regions, such as Xi’an Taxi, but as the number of regions increases, its top-k strategy struggles to capture complex correlations, resulting in performance degradation.

Compared to the above baselines, our model takes into account the high-level correlations in both spatial and temporal aspects of spatio-temporal data, leading to a more comprehensive modeling of spatio-temporal correlations. In addition, by incorporating the differential attention mechanism, we selectively amplify the most informative correlations while suppressing noise. These factors contribute to the superior performance of our model.

Beyond that, our method's extensive capture of high-level correlations might intuitively suggest a slower runtime. In PDFormer, this approach achieves a balance between performance and efficiency, excelling in both aspects. Therefore, we compared the runtimes of the two methods through multi-step forecasting experiments. As shown in~\autoref{tab:runtime_table}, the results demonstrate that our model is fast in both training and inference due to Flash-Attn support, validating our model's practical value.

\subsection{Ablation Study}
\begin{figure}
    \centering
    \includegraphics[width=\linewidth]{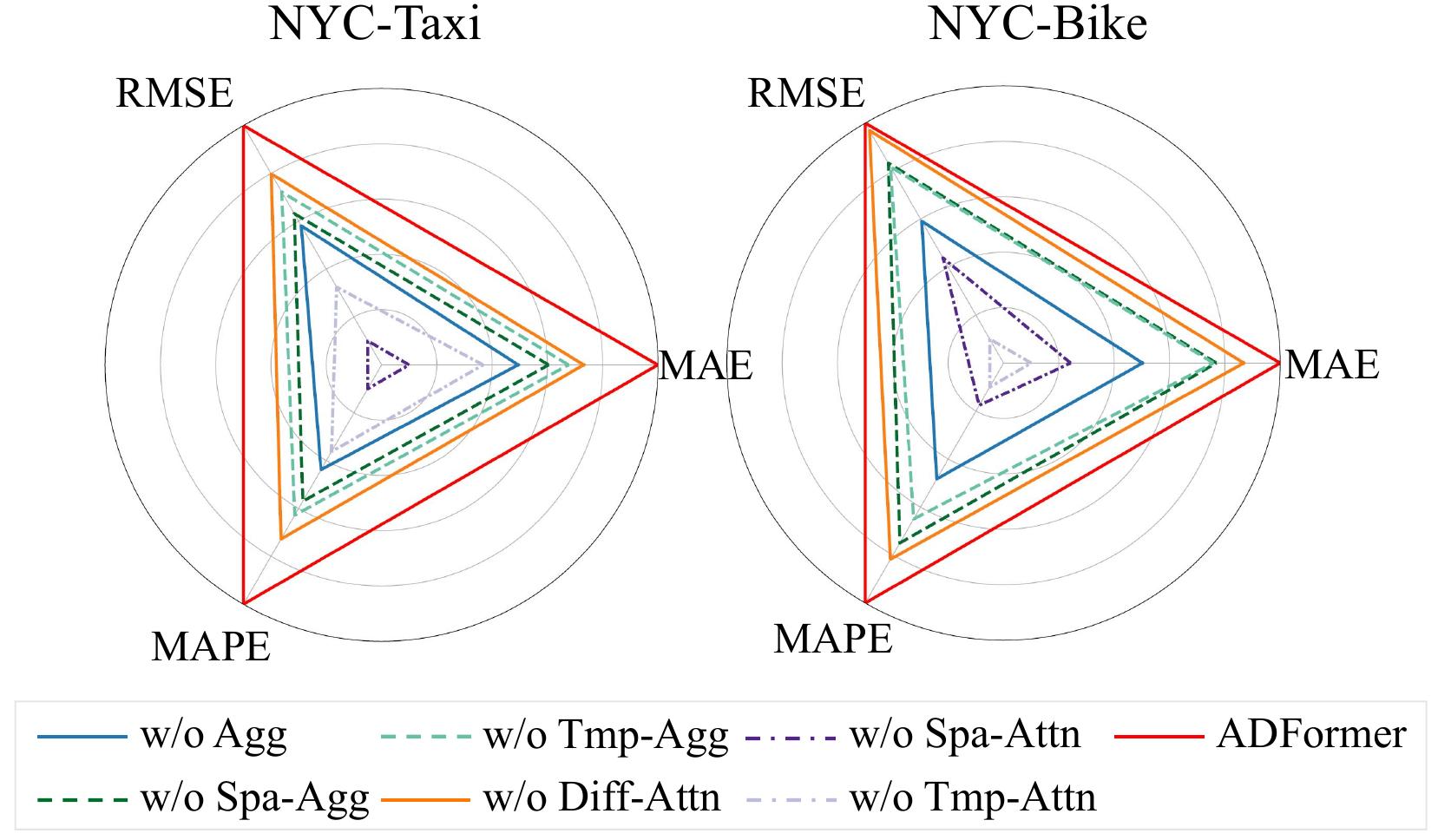}
    \caption{Ablation study on NYC datasets.}
    \label{fig:ablation}
    \vspace{-0.3cm}
\end{figure}

\begin{table}

    \centering
    \renewcommand{\arraystretch}{0.6}
    \resizebox{0.8\columnwidth}{!}{
    \tiny
    \begin{tabular}{c|l|r|r}
        \toprule
         Dataset & Phase & PDFormer & ADFormer \\
         \midrule
         \multirow{2}{*}{NYC-Taxi}
            & training & 53.75 & 35.51 \\
            & inference & 2.89 & 1.68 \\
        \midrule
         \multirow{2}{*}{NYC-Bike}
            & training & 48.17 & 25.62 \\
            & inference & 2.12 & 1.18 \\
        \bottomrule
    \end{tabular}%
    }
    \caption{Runtime (seconds) per epoch of two methods.}
    \label{tab:runtime_table}
    \vspace{-0.4cm}
\end{table}
We further validate the role of each module in the model by removing certain modules from our model.
(1) \textbf{w/o Diff-Attn} removes the differential attention mechanism in spatial correlations modeling.
(2) \textbf{w/o Agg} ignores the high-level spatio-temporal correlations, with the model only capturing correlations within the data at the original level.
(3) \textbf{w/o Spa-Agg} removes spatial aggregation and focuses solely on temporal aggregation.
(4) \textbf{w/o Tmp-Agg} considers only spatial aggregation.
(5) \textbf{w/o Spa-Attn} only captures correlations in the temporal aspect.
(6) \textbf{w/o Tmp-Attn} only captures correlations in spatial aspect.
We conduct multi-step prediction experiments on the NYC datasets. The experimental results are inverted and normalized to facilitate visualization, as shown in \autoref{fig:ablation}. In the figure, lines closer to the outer edges indicate better performance. We draw the following conclusions: 

(1) Capturing correlations from only spatial or temporal perspective is insufficient. As indicated by the dark and light purple dashed lines in the figure, spatial correlations are indispensable for taxi demand, while temporal correlations are pivotal for bike demand. This may be because taxi rides generally cover a larger spatial range and are more closely related to the functionality of dissimilar regions, whereas bike usage is more time sensitive. 

(2) Spatio-temporal aggregation is essential for modeling data correlations, as relying solely on temporal or spatial aggregation is limited. As shown by the dark and light green dashed lines, the relative importance of each varies: spatial aggregation is more impactful in the taxi dataset, while in the bike dataset, temporal aggregation plays a greater role. This aligns with the pattern observed at the first point. 

(3) The Differential Attention mechanism is particularly effective in spatially dominant scenarios like taxi demand, but less effective when temporal dependencies, such as those in bike demand, play a more significant role. This is mainly because it is applied only along the spatial dimension, given the relatively short temporal sequence length in our setting.

\subsection{Parameter Study}
To investigate the impact of the number of spatial clusters on the experimental results, we design a parameter study from the perspectives of both hierarchy and quantity on the Xi'an-Taxi dataset for single-step and multi-step forecasting tasks, while NYC datasets yield similar results. We first test a spatial aggregation-free architecture, followed by architectures with one-level, two-level, and three-level spatial aggregation, as shown in~\autoref{tab:parameter_study}. The results lead to two key conclusions:

(1) Forecasting accuracy improves regardless of the number of aggregation levels. And performance tends to enhance as the number of levels increases.

(2) The benefit of deeper aggregation levels becomes more pronounced at longer forecasting horizons, as high-level spatial correlations generally unfold over extended time spans, consistent with the temporal dynamics of human mobility between functional zones.

Notably, the results suggest that our model exhibits robustness with respect to this parameter. In practice, prior knowledge of functional zone categorizations within a city or specific regions can provide valuable guidance for parameter configuration.

\subsection{Case Study}
\begin{figure}
    \centering
    \includegraphics[width=0.9\linewidth]{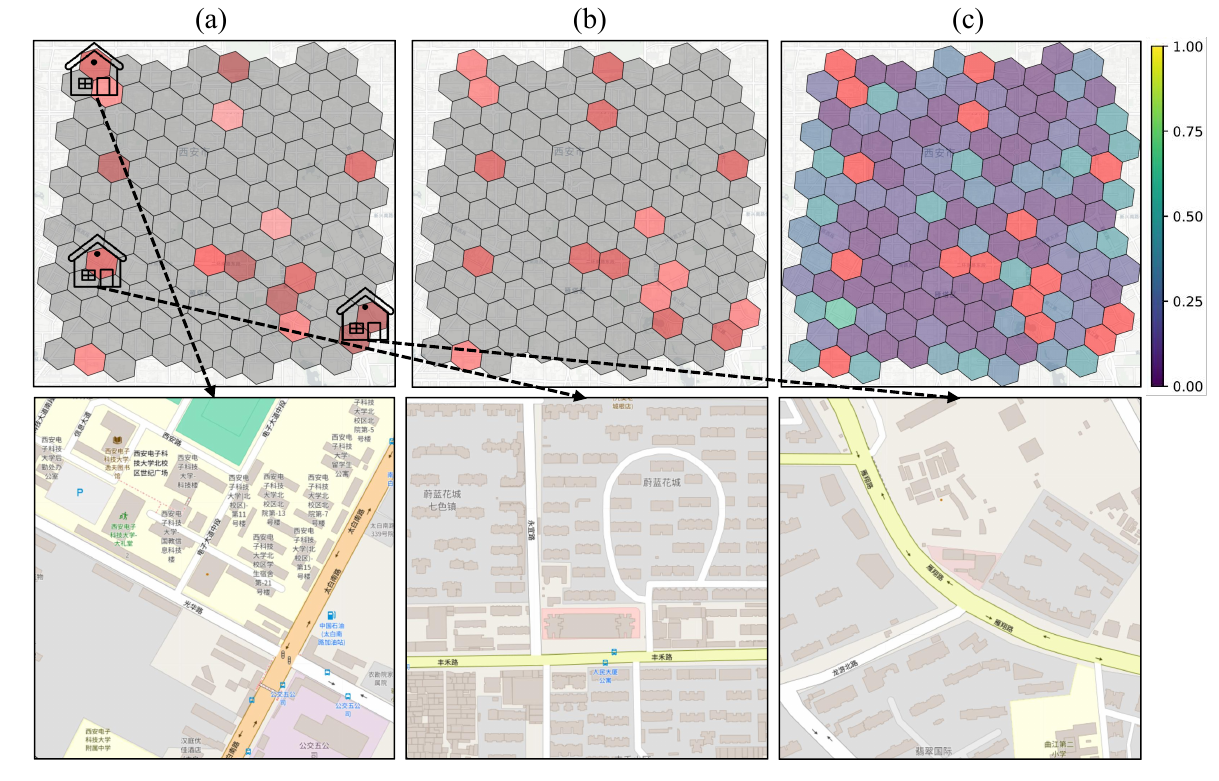}
    \caption{Spatial aggregation and high-level correlations.}
    \label{fig:case_study}
    \vspace{-0.3cm}
\end{figure}

\begin{table}[]
    \centering
    \resizebox{1.0\columnwidth}{!}{
    \begin{tabular}{c|c|l|rrrrrr}
         \toprule
         & Horizon & Metric & w/o spa-agg & [16] & [64] & [64,16] & [64,8] & [96,16,8] \\ 
        \midrule
        \multirow{6}{*}{\shortstack{Xi'an\\Taxi}}
            & \multirow{3}{*}{90 min} & MAE  & 3.782      & 3.713      & 3.719      & 3.697       & 3.705        & 3.694        \\
            &                         & RMSE & 5.601      & 5.488      & 5.511      &  5.466      & 5.488        & 5.457        \\
            &                         & MAPE & 22.824      & 22.525      & 22.540      & 22.465       & 22.454        & 22.402        \\ \cmidrule{2-9}
            & \multirow{3}{*}{3 hour} & MAE  & 3.935      & 3.875      &  3.849     & 3.830       &  3.836       & 3.814        \\
            &                         & RMSE & 5.880      &  5.786     & 5.744      &  5.706      & 5.727        & 5.689        \\ 
            &                         & MAPE & 23.658      &  23.291     & 23.147      &  23.114      & 23.093        & 23.039        \\
        \bottomrule
    \end{tabular}
    }
    \caption{Impact of spatial cluster numbers and hierarchical levels.}
    \label{tab:parameter_study}
    \vspace{-0.2cm}
\end{table}
We visualize the high-level spatial correlations learned by our model to better illustrate the effectiveness of our approach. As illustrated in~\autoref{fig:case_study}, the urban area in the Xi'an dataset is divided into hexagonal grids. In (a), there are six clusters filled with dark or light red. In reality, these grids mostly contain residential areas. The second row of maps displays three grids that are relatively similar. These grids not only contain a large number of residential areas, but also include major intersections, indicating their spatial similarity and the potential for aggregation. Subfigure (b) shows larger clusters that are formed by aggregating the smaller clusters in (a), incorporating more regions. Finally, subfigure (c) presents the aggregation results at the highest level and illustrates the relationships between clusters, where grids of the same color belong to the same cluster.
The core idea of spatial aggregation is to break artificially defined spatial boundaries and view spatial correlations from a holistic perspective. The concept of temporal aggregation is similar to this.
\section{Related Works}

Early works~\cite{zhang2016dnn,yao2018deep,ke2018hexagon} widely adopted Convolutional Neural Networks (CNNs), modeling urban regions as grids with each area represented as a pixel, and using CNNs to capture spatio-temporal dependencies.
However, geospatial data has inherent topological characteristics, such as spatial irregularity and local perception, which make Graph Neural Networks~(GNNs)
more suitable than CNNs for modeling these structures. For example, 
~\cite{geng2019spatiotemporal} integrates non-Euclidean pairwise correlations among regions through multi-graph convolution.
~\cite{pian2020spatial} uses an attention mechanism to assign importance to neighboring regions and builds a dynamic graph to capture time-specific spatial relationships.
~\cite{jin2022deep} integrates 1D CNNs and multi-graph attention networks to capture short-term spatial and long-term temporal dynamics.
While graph structures intuitively model spatial connections, defining adjacency matrices that accurately reflect spatial dynamics remains challenging.

To better tackle this issue, several attention-based spatio-temporal prediction models have been proposed. These models aim to learn spatio-temporal dynamics directly from the data rather than pre-defining the graph structure. For instance,~\cite{yan2021learning} uses a K-hop adjacent mask to allow the model to focus on local information.~\cite{jiang2023pdformer} incorporates two masking matrices in spatial self-attention to capture both short-range spatial correlations based on distance and long-range correlations based on historical data similarity.~\cite{liang2024asstformer} applies a self-corrected top-K mechanism to filter attention scores, achieving attention denoising.
Meanwhile, recent methods focus on processing input data, such as incorporating adaptive embeddings and decomposing from the perspective of channel or variables~\cite{liu2023spatio}~\cite{qiu2025duet}~\cite{wu2024catch}~\cite{qiu2025comprehensive}.
However, these models often overemphasize correlations between unrelated objects, limiting their ability to focus on more relevant spatio-temporal relations. In this paper, we propose a method to reduce attention noise from unrelated regions while reinforcing the connections between related regions and consider holistic spatio-temporal correlations, thereby improving spatio-temporal forecasting accuracy.
\vspace{-0.2cm}

\section{Conclusion}
In this work, we propose the Aggregation Differential Transformer (ADFormer) for passenger demand forecasting. ADFormer integrates Differential Attention to denoise raw spatial correlations and adopts a region aggregation strategy to capture high-level, non-pairwise spatial dependencies. Additionally, a hierarchical temporal matrix approach is developed to model broader and more intricate temporal dependencies. The effectiveness of ADFormer is evaluated on public datasets, including ablation studies that validate the contributions of the aggregation strategies and Differential Attention. Our findings reveal that the influence of temporal and spatial correlations varies across different scenarios, reflecting the inherent nature of the datasets. Furthermore, the parameter study and the case study quantitatively and qualitatively demonstrate the effectiveness of spatial aggregation.

\section*{Acknowledgements}
This work was partially supported by National Natural Science Foundation of China (62472174). This work was supported by “the Open Research Fund of Key Laboratory of Advanced Theory and Application in Statistics and Data Science–MOE,ECNU”. Jilin Hu is the corresponding author of the work.


\bibliographystyle{named}
\bibliography{ijcai25}

\end{document}